\documentclass{article}

\usepackage[preprint]{neurips_2022}

\usepackage[utf8]{inputenc} 
\usepackage[T1]{fontenc}    
\usepackage{hyperref}       
\usepackage{url}            
\usepackage{booktabs}       
\usepackage{amsfonts}       
\usepackage{nicefrac}       
\usepackage{microtype}      
\usepackage{xcolor}         
\usepackage{algorithmic}    
\usepackage[ruled,vlined]{algorithm2e}  
\usepackage{multicol} 
\usepackage{multirow} 
\usepackage{bigstrut} 
\usepackage{wrapfig}
\usepackage{graphicx}
\usepackage{float}
\usepackage{subcaption}
\usepackage{textpos}
\usepackage{mwe}

\newcommand\braces[1]{\left\{#1\right\}} 
\newcommand\parens[1]{\left(#1\right)} 

\title{Deep Neural Imputation:  A Framework for Recovering Incomplete  Brain Recordings}

\author{%
  Sabera Talukder\thanks{Equal contribution.}\\
  Caltech\\
  \texttt{sabera@caltech.edu}
  \And
  Jennifer J. Sun\footnote[1]{}\\
  Caltech\\
  \texttt{jjsun@caltech.edu}
   \AND
   Matthew Leonard \\
   UCSF\\
   \texttt{Matthew.Leonard@ucsf.edu} \\
   \And
   Bingni W. Brunton \\
   U of Washington \\
   \texttt{bbrunton@uw.edu} \\
   \And
   Yisong Yue \\
   Caltech \\
   \texttt{yyue@caltech.edu} \\
}

\begin{document}

\maketitle

\begin{abstract}

Neuroscientists and neuroengineers have long relied on multielectrode neural recordings to study the brain. However, in a typical experiment, many factors corrupt neural recordings from individual electrodes, including electrical noise, movement artifacts, and faulty manufacturing. Currently, common practice is to discard these corrupted recordings, reducing already limited data that is difficult to collect. To address this challenge, we propose \textit{Deep Neural Imputation (DNI)}, a framework to recover missing values from electrodes by learning from data collected across spatial locations, days, and participants. We explore our framework with a linear nearest-neighbor approach and two deep generative autoencoders, demonstrating DNI's flexibility. One deep autoencoder models participants individually, while the other extends this architecture to model many participants jointly. We evaluate our models across 12 human participants implanted with multielectrode intracranial electrocorticography arrays; participants had no explicit task and behaved naturally across hundreds of recording hours. We show that DNI recovers not only time series but also frequency content, and further establish DNI's practical value by recovering significant performance on a scientifically-relevant downstream neural decoding task.

\end{abstract}

\section{Introduction}
Multielectrode recordings measure the dynamic activation of large networks of neurons in the brain and are a key enabling tool in studying neural function at scale \cite{paulk2022large, anumanchipalli2019speech, willett2021high, carmena2003learning}. 
Modern multielectrode recordings can monitor neural activity continuously in numerous brain regions with high temporal frequency across days. In humans, implanting such electrode arrays is an invasive clinical procedure performed by neurosurgeons. However, individual electrode implants can fail during any of the recording days, almost always leading to missing electrode values in these difficult-to-acquire datasets. The current standard practice is to discard electrodes with missing data, which excludes potentially useful information from analysis and also limits the generalizability of models across days and individuals~\cite{banga2022reproducibility, anumanchipalli2019speech, ruebel2019nwb}.

\begin{figure}
    \centering
        \includegraphics[width=\linewidth]{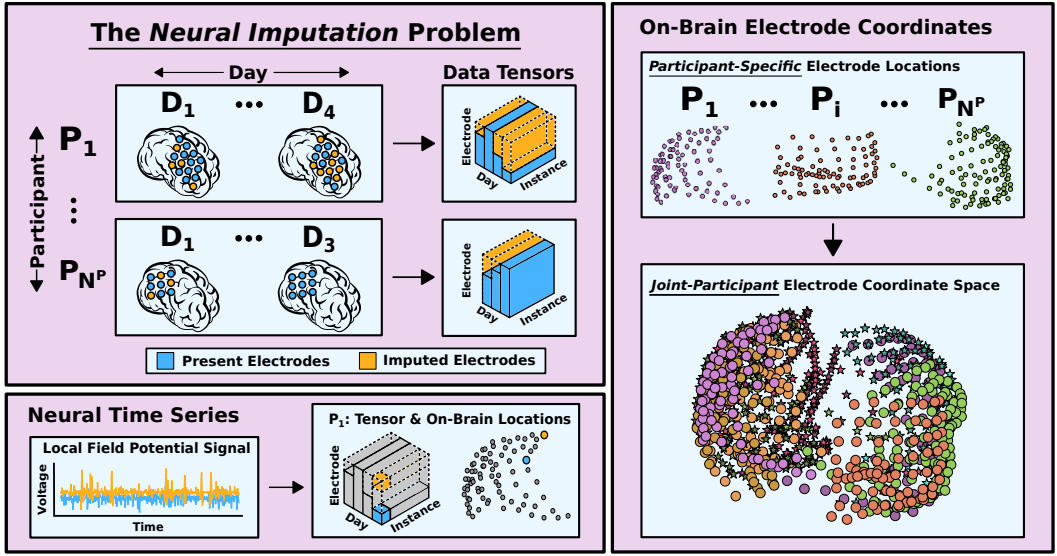}
        \vspace{-0.2in}
        \caption{\textbf{Overview of Neural Imputation}. Neural imputation recovers missing electrode signals (yellow) across days and participants given non-corrupted data (blue). \textit{Top left:} Observed multielectrode data from each participant is a ragged tensor, because participants have different sets of electrodes that may be missing in different days. In a typical neuroscience study, a few hours from a single day in a single participant are treated as an independent dataset for analysis (see~\cite{collinger2013high,anumanchipalli2019speech,natraj2022compartmentalized}, among many others).
        \textit{Bottom left:} A time series instance from a single electrode in a specific brain location measures a voltage trace across time. Depending on the specific electrode type, this trace may represent the electrical activity of a single neuron or the average of tens-of-thousands of neurons. In our dataset, we measure local field potential (LFP) signals. \textit{Right:} An additional challenge in imputation is variation in brain morphology and spatial electrode locations across participants.}
    \label{fig:intro_fig}
    \vspace{-0.5cm}
\end{figure}

To address this challenge, we study the problem of imputing missing values from multielectrode data using recordings collected across multiple days in several human participants (Figure~\ref{fig:intro_fig}). While related to the generic problem of time-series imputation (e.g., \cite{liu2019naomi,luo2018multivariate}, more discussion in Section~\ref{sec:related}), previous works are not directly applicable
here due to additional challenges associated with multielectrode recordings. Two major challenges are that (1) electrodes are often missing for entire sessions with no adjacent recorded timestamps, and (2) existing methods do not handle data across participants who have completely different sets of electrodes. To our knowledge, no prior work has conducted a rigorous study of how to recover or impute missing electrode recordings using data across sessions and different human participants. 

We propose \textit{Deep Neural Imputation (DNI)}, a framework to recover missing electrode data by learning across days and participants. We instantiate our framework with both a linear nearest-neighbor method as well as two deep autoencoder-based generative models. DNI uses a self-supervised task based on mask-filling during training, which we call masked electrode modeling, similar to previous masking approaches in other domains~\cite{devlin2018bert,liu2019naomi}. In order to leverage data across participants, we extend our learning approach to use participant-specific encoding/decoding layers along with shared inner layers to jointly model participants. 

We evaluate DNI's performance on multielectrode intracranial electrocorticography recordings from 12 human participants with naturalistic behavior across multiple days~\cite{peterson2022ajile12}. Because of the large inter-participant and inter-day variability\cite{gonschorek2021removing},\footnote{Spatial sampling in intracranial datasets is clinically determined and therefore highly variable. Consequently, recovering missing data is far more challenging than, e.g., EEG data where spatial sampling is easily controlled.} each participant's recordings over a day (and often only a few hours of said recordings) would typically be treated as separate datasets in neuroscience studies (e.g., \cite{collinger2013high,anumanchipalli2019speech,natraj2022compartmentalized}). Despite these variations, we show that our approach can recover electrode recordings even when a significant fraction is missing. 
Further, we explore DNI's ability to recover frequency content, which is significant to the neuroscience community \cite{miller2007spectral,jas2017learning, cole2019cycle}. We also apply our imputed data on a downstream neural decoding task ~\cite{peterson2021generalized}, showing that our method can recover significant decoding power when compared to the non-imputed missing data. These results suggest that our methods can help address a major challenge in experimental design and analysis of multielectrode arrays in neuroscience.

Our contributions are summarized as follows:

\begin{itemize}
\vspace{-0.1in}

\item We propose the Deep Neural Imputation (DNI) Framework, a method for recovering missing multivariate electrode time series recordings across sessions and human participants. To our knowledge, DNI is the first framework to simultaneously impute fully missing spatial and temporal time series from multielectrode recordings. 

\item We empirically demonstrate DNI's compatibility with both linear and nonlinear imputation methods, and generalizability to the joint-participant regime.

\item We provide experimental evidence of DNI's direct and immediate utility in downstream scientific analyses: (1) DNI reconstructs not only time series content but also frequency based power-spectral content, and (2) DNI's spatiotemporal reconstructions directly improve a brain decoder's classification accuracy when missing data is present.
\end{itemize}

\section{Background \& Problem Setup}
\textbf{Data Modality \& Format.}
Our data, further described in Section \ref{sec:exp_data}, consists of participants (indexed by $i$), recording days (indexed by $j$),\footnote{Most often, a neuroscience recording session is less than or equal to a few hours. However, for simplicity of exposition we use 24-hr days since that is how our data is organized.} electrodes corresponding to spatial locations (indexed by $k$), and time series instances (indexed by $t$). In other words, an observation $E_{i,j,k,t}$ corresponds to time series instance $t$ from electrode $k$ on day $j$ for participant $i$. Indices that follow others demonstrate dependence; for example, time series instance $t$ depends on the specific electrode $k$, the specific day $j$, and the specific participant $i$. Here we define generalizable as being able to perform on previously unseen data. When our models impute data across dimensions $j,k,t$ they generalize across time series instances, spatial locations, and days (Figure~\ref{fig:intro_fig}) and we call them \textit{day-generalizable}. When our models impute data across dimension $i$ as well as dimensions $j,k,t$ they additionally share a joint participant embedding space; we call them \textbf{\underline{D}}ay-with-j\textbf{\underline{O}}int-\textbf{\underline{P}}articipant-\textbf{\underline{E}}mbeddings-generalizable, or \textit{DOPE-generalizable}. By necessity, DOPE-generalizable models are also day-generalizable.

\textbf{Challenges in Modeling Electrode Recordings}. Due to significant inter-participant and inter-day variability, data analysis methods in neuroscience commonly treat a few hours from single participants as individual datasets \cite{collinger2013high,anumanchipalli2019speech,natraj2022compartmentalized} to be analyzed separately. Figure \ref{fig:intro_fig}'s right panel demonstrates the stark electrode configuration variability in our participants, illustrating a common phenomenon in this data type. Further sources of this variability include movement artifacts, insurmountable experimental noise, and electrode recording failure. For example, in a neuropixel single-unit electrophysiology mouse study, only 55.4\% of the data could be used for downstream analyses, because of ``recording failure[s],'' ``low yield,'' and ``noise/artifact[s]'' \cite{banga2022reproducibility}. In this ECoG LFP human speech decoding study, electrode recordings were discarded from a majority of the participants due to ``bad signal quality'' \cite{anumanchipalli2019speech}. The pervasiveness of missing data is exemplified by the standard data storage file format in neuroscience, Neurodata Without Borders, which explicitly emphasizes its support for ``dense ragged arrays'' that permit ``missing fields'' \cite{ruebel2019nwb}.

These ragged data tensors result from the lack of consistency across the dimensions $i,j,k,t$  (Top Left Figure \ref{fig:intro_fig}). Since ragged data are difficult to jointly analyze, they are most commonly broken up into smaller, non-ragged tensors. This procedure leads to more manageable data for neuroscience analysis pipelines, meaning that discarding data remains a common practice despite its inefficiency. Because neuroscience data processing pipelines seldom handle variation in the expected data structure, these pipelines fundamentally limit the generalization capability of neuroscience models.

\textbf{Formal Imputation Goals.} To address the above challenges, our goal is to impute missing electrode values by learning from multielectrode time series data across spatial locations, days, and participants.

Let $\mathcal{E}_{i,j} = \{{E_{i,j,k}} \mid k \in \mathcal{K}_i\}$ be the full set of active electrodes on day $j$ for participant $i$. $\mathcal{K}_i$ represents the full set of electrodes for participant $i$, which remains consistent across days. Let $\mathcal{M}_{i,j} \subset \mathcal{K}_i$ represent the set of missing electrodes on day $j$ for participant $i$. Let $\bar{\mathcal{E}}_{i,j} = \{E_{i,j,k} \mid k \in \mathcal{K}_i \setminus \mathcal{M}_{i,j}\}$ represent the set of observed electrodes. 

Formally, DNI seeks to recover $\mathcal{E}_{i,j}$ given $\bar{\mathcal{E}}_{i,j}$. Previous work for multivariate time series imputation~\cite{liu2019naomi,cao2018brits,luo2018multivariate} studies missing data at the level of sequences $\braces{t_1, t_2, ..., t_T}$, where there may be missing observations for a set of timestamps (e.g. $\braces{t_5, t_6, t_7}$). However, these methods could not impute when a feature was missing for all timestamps (e.g. $\braces{t_1, t_2, ..., t_T}$); in other words, each feature required at least one observed timestamp. Therefore, these methods are not directly applicable to ours because an entire time series instance $E_{i,j,k,t} = \braces{t_1, t_2, ..., t_T}$ may be missing and have no adjacent timestamps.

\section{Deep Neural Imputation Framework}
\begin{figure}
    \centering
        \includegraphics[width=\linewidth]{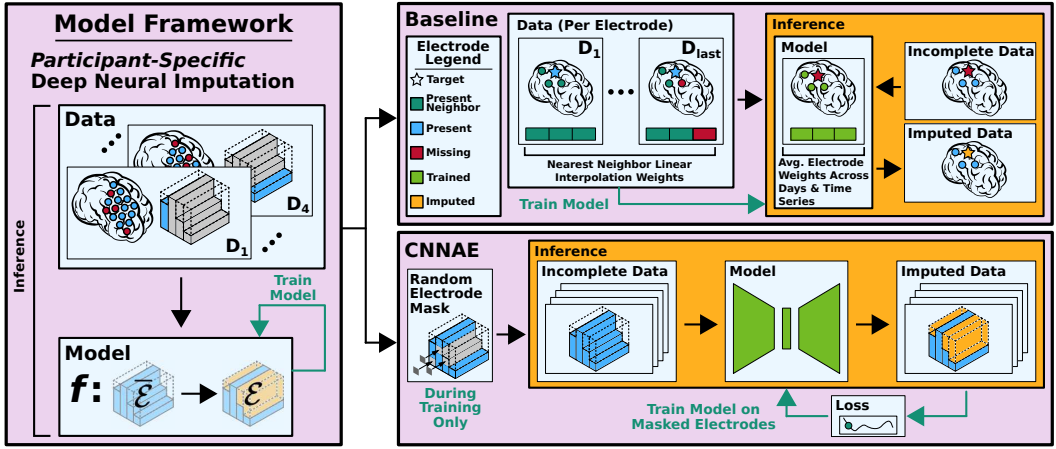}
        \caption{\textbf{Deep Neural Imputation with Baseline and CNNAE Instantiations}. \textit{Left:} In the participant-specific setting, we impute missing electrode recordings using data across days from the same participant. The goal is to map observed electrodes $\bar{\mathcal{E}}$ to the full set $\mathcal{E}$. \textit{Top Right:} DNI instantiation of the linear nearest neighbor baseline.
        \textit{Bottom Right:} DNI instantiation of the CNNAE trained with masked electrode modeling.}
    \label{fig:model}
    \vspace{-0.5cm}
\end{figure}

\begin{wrapfigure}{r}{0.48\textwidth}
\vspace{-0.2in}
    \centering
        \includegraphics[width=\linewidth]{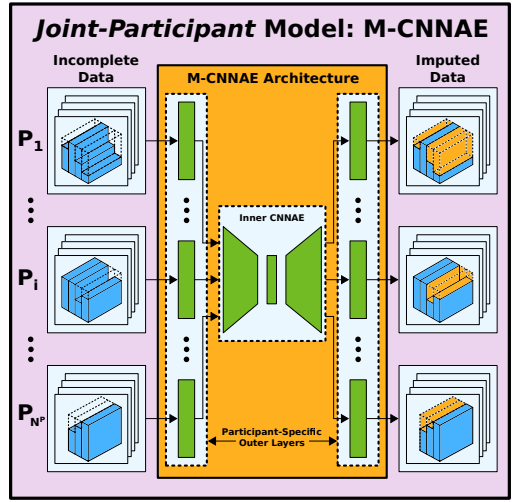}
  \vspace{-0.2in}
    \caption{\textbf{Joint-Participant Deep Neural Imputation}. We extend our framework to jointly model many human participants, with different electrode spatial configurations, using participant-specific encoding and decoding layers and a shared CNNAE backbone.
    }
    \label{fig:patient_model}
\vspace{-0.35in}
\end{wrapfigure}

The goal of Deep Neural Imputation is to learn an imputation function $f: \bar{\mathcal{E}}_{i,j} \rightarrow \mathcal{E}_{i,j}$, as depicted in Figure~\ref{fig:model} (Left). $f$ may be either linear or nonlinear, and we study both instantiations. In the participant-specific case, we learn a function $f^{(i)}$ for each participant $i$ (Figure~\ref{fig:model}). In the joint-participant case, we learn a single function $f$ across all participants (Figure~\ref{fig:patient_model}). In all cases, our learned functions do not vary by recording day, making our models day-generalizable.

\textbf{High-Level Approach.}
To our knowledge, no prior work has rigorously studied neural electrode signal imputation, therefore one of our key design goals is to identify the simplest approach that works well on this challenging imputation task. Two salient desiderata for imputation methods compatible with DNI are (1) the recovery of missing data without adjacent timestamps using observations from different days; and (2) robustness to variations in brain morphology and physical electrode placement across participants. Participant-specific modeling addresses (1) but not (2) and is day-generalizable, while joint-participant modeling addresses both (1) and (2) and is DOPE-generalizable.

To create day-generalizable models that address (1), the key idea is to learn conserved relationships between electrodes across recording days. A conceptually straightforward method is  nearest neighbors linear interpolation, which we propose as a natural baseline in our framework since no other baselines exist. We also study a nonlinear deep autoencoder model, which is discussed below. We then extend the aforementioned participant-specific autoencoder model to train jointly over all participants. This DOPE-generalizable model additionally addresses (2).

\textbf{Linear Baseline}. In the absence of appropriate existing baselines, we looked to known properties of neural signals for inspiration. Correlations exist across spatially close electrode neighbors; this well known observation motivated us to impute missing electrodes with linear combinations of observed, neighboring electrodes. In particular, we compute time series correlations from the $N$ nearest neighbors of an electrode on observed days (e.g. ``training days''). Then, we use these correlations as weights to linearly combine the time series from missing electrode neighbors on the held out ``test day.'' These weighted, linearly combined time series compilations constitute our reconstructed neural signal and form our day-generalizable baseline model.

\textbf{CNNAE}. In our Convolutional Neural Network AutoEncoder (CNNAE) learning approach (Figure~\ref{fig:model} Bottom Right), the encoder $f_{\theta}^{(i)}$ maps $\bar{\mathcal{E}}_{i,j}$ with zero-filled missing electrodes to an embedded representation $z_{i,j}$. The decoder $f_{\psi}^{(i)}$ then maps $z_{i,j}$ to $\hat{\mathcal{E}}_{i,j}$, which is either a reconstruction or an imputation (which we distinguish below) of the full set of electrodes $\mathcal{E}_{i,j}$. For $k \in \mathcal{K}_i \setminus \mathcal{M}_{i,j} = \overline{\mathcal{E}}_{i,j}$, $\hat{E}_{i,j,k}$ corresponds to a \textit{reconstruction} of \textit{observed} electrode data. For $k \in \mathcal{M}_{i,j}$, $\hat{E}_{i,j,k}$ corresponds to an \textit{imputation} of \textit{unobserved} electrode data. Imputation is a more challenging task than reconstruction, since in imputation, the decoding target is not an input to the model. The CNNAE is trained using the self-supervised objective described in Section~\ref{sec:objective}.  Specific architecture design choices are discussed in Section \ref{sec:exp_train} -- the key criterion is to effectively encode and decode multivariate electrode time series data. The CNNAE is a day-generalizable model, but it forms the backbone for the DOPE-generalizable model class described below.

\textbf{M-CNNAE}. For joint-participant imputation, the Multihead CNNAE model (M-CNNAE) learns a single function $f$ for all participants, instead of one $f^{(i)}$ for each participant. When extending the CNNAE to jointly model different participants (Figure~\ref{fig:patient_model}), we first used a participant-specific encoding layer to map input electrodes to a shared embedding space. This shared space is necessary because each participant has a different number of electrodes arranged in highly varied configurations. The encoder, $f_\theta$, and decoder, $f_\psi$, are then trained using the CNNAE's learning procedure. To map the representation from the shared embedding space back to the participant-specific electrode configuration, we also train a participant-specific decoding layer.
Components are trained jointly with data from all participants via a self-supervised objective (Section~\ref{sec:objective}). Unlike the CNNAE, the M-CNNAE trains a shared representation across all participants, making it DOPE-generalizable.

\subsection{Training Objective}
\label{sec:objective}

We train our CNNAE and M-CNNAE models with self-supervision using masked electrode modeling, where we randomly mask $l$ observed electrodes during training. Let $\tilde{\mathcal{E}}_{i,j}$ represent the observed electrodes with random masking. Then, we train the autoencoder to simultaneously impute the masked values as well as reconstruct the observed values. This task is similar to masking used in previous time series imputation approaches~\cite{liu2019naomi}, except we mask entire time series from an electrode, instead of a subset of timestamps. The autoencoder is trained with the following objective: 
$\mathcal{L}_{i,j}^{\it{NLL}} = -\log\parens{p_{f_{\psi}^{(i)}}\parens{\bar{\mathcal{E}}_{i,j}|f_{\theta}^{(i)}(\tilde{\mathcal{E}}_{i,j})}}.$

\section{Experiments}
We study Deep Neural Imputation with a linear baseline model, CNNAE, and M-CNNAE on real-world multielectrode recordings from 12 human participants performing naturalistic behavior across multiple recording days. In Section~\ref{sec:exp_data} we further describe the dataset, in Section~\ref{sec:exp_train} we expand upon our training and evaluation setups, and in Section~\ref{sec:exp1} we compare our three methods. We then demonstrate DNI's value to computational neuroscience by showing the frequency content learned by our method in Section~\ref{sec:exp2}, and recovering significant neural decoding performance with our imputations when missing values are present in Section~\ref{sec:exp3}.

\subsection{Datasets for Deep Neural Imputation}~\label{sec:exp_data} To study Deep Neural Imputation we utilize recently released data from all 12 AJILE12~\cite{peterson2022ajile12} participants, each with $\sim$100 electrodes recorded across multiple days. Each participant has a wide range of naturalistic behaviors during recording, making reconstruction/imputation more difficult than if explored with canonical task-based neural data \cite{peterson2021behavioral}. In the past, deep learning models have been used, often on subsets of the AJILE12 participants. However, these models have explored more classical neural decoding tasks, namely binary prediction and binary classification of arm movement~\cite{wang2018ajile, peterson2021generalized, peterson2021learning}.

\textbf{Data Processing Pipelines}. Using the same training split as \cite{peterson2021generalized} where the last day is held out as the test day, we define two different data processing pipelines: 
\vspace{-0.1in}
\begin{itemize}
    \item Procedure A: lower frequency content data used in Section~\ref{sec:exp1}
    \vspace{-0.02in}
    \item Procedure B: higher frequency content data used in
    Sections~\ref{sec:exp2},~\ref{sec:exp3}
        \vspace{-0.02in}
\end{itemize} 

We use Procedure A to study our models' ability to reconstruct time series data. We perform standardization on a 50,000 sample segment, corresponding to a 100$sec$ length time series, on a per electrode, per day, per participant basis. We divide the 100$sec$ into 20$sec$ and 80$sec$ sections, calculate the mean ($\mu_{20}$) and standard deviation ($\sigma_{20}$) of the 20$sec$ segment, and for each time step in the 80$sec$ segment compute $\frac{Time Step - \mu_{20}}{\sigma_{20}}$.  As typically done in neuroscience studies without task labels, performing local standardization on each time segment allows us to account for neural data distribution changes such as spatial electrode shifts, neural drift, habituation to external stimuli, etc. We then downsample this data by a factor of 100, resulting in 400 time steps of data at a frequency of $5Hz$.

Using Procedure B, we study the performance of our imputation model with two downstream tasks: frequency content preservation, and a scientifically-relevant neural decoding task. We follow the data processing pipeline outlined in \cite{peterson2021generalized}, consisting of band-pass filtering, downsampling, and trimming, resulting in $\sim$1000 time steps of data at a frequency of 250$Hz$.

Our main goal with DNI is to reconstruct the neural time series; therefore, with Procedure A we aggressively downsampled to reduce the high temporal resolution of the original data. When using Procedure A, the frequency content of the reconstructions was recovered in the limited frequency bands available for analysis (i.e. lower frequency bands due to downsampling). The purpose of Procedure B was to (1) test our models in a regime that featured different time series lengths and frequency content in the training data, (2) verify the frequency content recovery property in higher frequency bands, (3) use our reconstructions in a downstream neural decoding task.

\subsection{Training and Evaluation Setup} 
\label{sec:exp_train}

Our train and test split follows~\cite{peterson2021generalized}, where each participant's last day is held out as the test day. For the linear baseline model, we use 3 nearest neighbors to compute the correlation weights on the train days. We then apply these weights to the time series from the 3 nearest neighbors on the test day and combined them to reconstruct/impute the missing electrode's neural signal. For the CNNAE and M-CNNAE, we built the encoder with strided temporal convolutions and we based the decoder on~\cite{oord2016wavenet}; together these comprise the shared backbone architecture (see Appendix for more modeling details). During the CNNAE and M-CNNAE training, we use our self-supervised training objective to perform masked electrode modeling. Masked electrode modeling (Figure ~\ref{fig:model}) consists of creating a random electrode mask where $5$-$10 \%$ of the electrodes in each batch update are zero-filled. We then train the models to decode the full set of input electrodes; decoded electrode values are either reconstructed or imputed.

For evaluation, we use Pearson's Correlation between the original time series and reconstructed/imputed time series. We explore our model in 3 missing data regimes: $10\%$ missing, $20\%$ missing, and $50\%$ missing. Within each missing data regime, we average our correlation results over all time series instances, electrodes, and 3 distinct randomly generated sets of missing electrodes. A crucial step in evaluating our models' decoding capabilities is comparing our reconstructions/imputations against the ground truth data that is unseen by the model. It is worth noting that the 12 participants' original data does have naturally missing electrodes. In these cases we cannot compare our imputations because ground truth does not exist. This motivated us to create zero-filled masked electrode modeling because it allows us to evaluate not only our decoded reconstructions, but also our decoded imputations.

\begin{figure}
    \centering
        \includegraphics[width=\linewidth]{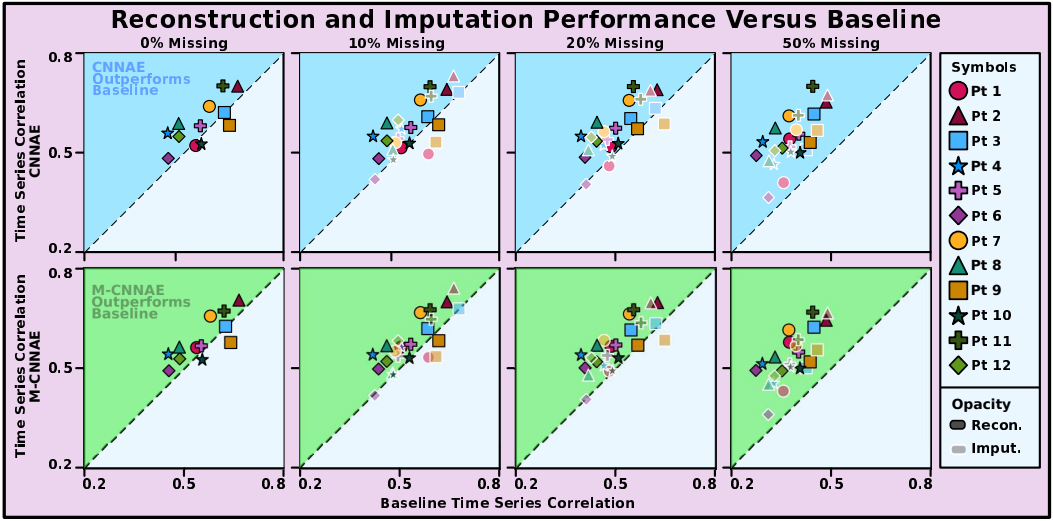}
        \vspace{-0.2in}
        \caption{\textbf{Comparing Time Series Correlation Across Deep Neural Imputation Methods}. The top row compares the CNNAE, and the bottom row compares the M-CNNAE to the Linear Nearest Neighbor Baseline respectively. The shaded triangle above the diagonal shows where the CNNAE and M-CNNAE respectively outperform the Baseline. Each point corresponds to a evaluation setting for one participant; means of 9 runs are shown (3 runs of each model with 3 sets of missing electrodes). Note that virtually all prior methods ignore missing data, therefore the conventional baseline approach would get a time series correlation of 0.  }\label{fig:dni_res}
        \vspace{-0.25cm}
\end{figure}

\subsection{Benchmark Results}\label{sec:exp1}
All time series correlation values report the comparison of our reconstruction/imputation with the original time series. It is worth re-emphasizing that imputation is a harder task than reconstruction, because for reconstruction the model sees the input reconstruction targets, while for imputation, the model only sees zero-filled time series values. In Figure~\ref{fig:dni_res} we compare the time series correlations values between the CNNAE and our baseline, top row, as well as the M-CNNAE and our baseline, bottom row. We note that prior approaches for imputing missing data in the neuroscience community use zero-filling, which results in a correlation value of zero.

We find that Deep Neural Imputation's linear baseline performs moderately well, validating our intuition behind shared information between neighboring electrodes. As the percentage of missing electrodes increases, the baseline performance understandably falls because there are fewer neighboring electrodes available. At the same time, the performance of both the CNNAE and M-CNNAE holds across our missing data regimes and improves over the baseline. For $0\%$ missing data the CNNAE outperforms the baseline for 9 participants, and in the most challenging evaluation regime ($50\%$ missing data) the CNNAE outperforms the baseline for all 12 participants in both reconstruction and imputation. The M-CNNAE model shares the same trend as the CNNAE when compared to the baseline, and in the most difficult evaluation regime ($50\%$ missing data) also outperforms the baseline for all 12 participants in both reconstruction and imputation. 

When studying the differences between the M-CNNAE and CNNAE, we found that participant 1 had the greatest performance improvement when using the M-CNNAE. Despite the fact that there are 12 trained CNNAE models (one for each participant) and 1 jointly-trained M-CNNAE model, there are more cases where the M-CNNAE to baseline performance is better than CNNAE to baseline performance. To decipher the M-CNNAE and CNNAE differences we decided to explore the M-CNNAE joint-participant representation space and CNNAE's participant-specific representation spaces. For a similar comparison, we . For a similar comparison, we stacked all 12 of the CNNAE's participant-specific representation spaces and analyzed the concatenation. The M-CNNAE's joint-participant embedding had more samples mapped to a shared space, when compared to the CNNAE's concatenated embedding space which had far more participant-specific clusters.

\subsection{Frequency Correlation Analysis}\label{sec:exp2}
\begin{figure}
    \centering
        \includegraphics[width=\linewidth]{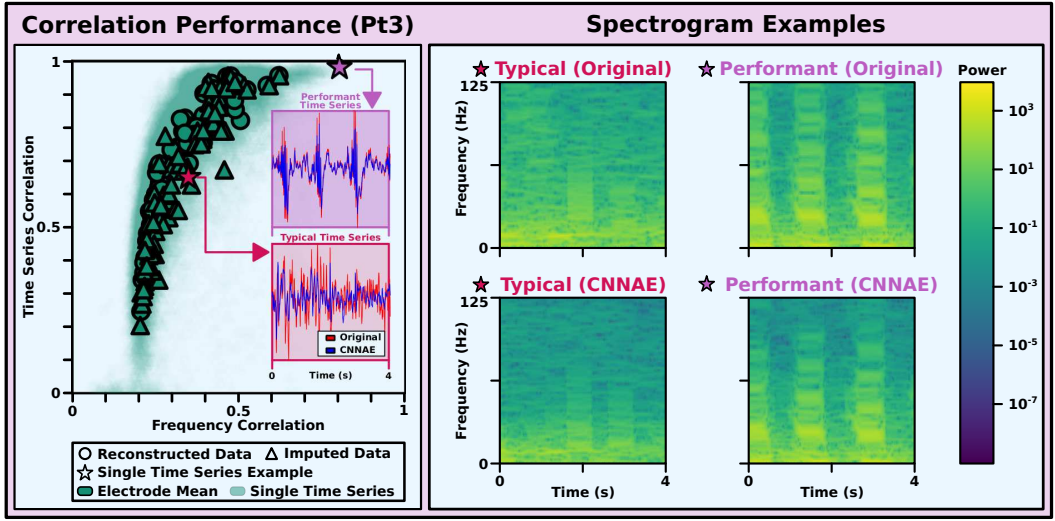}
        \vspace{-0.2in}
        \caption{\textbf{Relationship between Frequency and Time Series Correlations}. \textit{Left:} For participant 3 we plot the frequency vs. time series correlations along with two time series examples. Results for all patients can be found in the Appendix. \textit{Right:} Spectrogram examples corresponding to the time series examples on the left for a typical example and a performant example.}
    \label{fig:freq_corr}
    \vspace{-0.25cm}
\end{figure}

In Figure~\ref{fig:freq_corr}, we explore the relationship between frequency correlation and time series correlation across two proportions of missing data ($0\%$ \& $50\%$) because of frequency content's significance to the neuroscience community \cite{jas2017learning, cole2019cycle}. We observe a positive relationship between these correlations, despite the fact that our models were not trained to perform reconstruction or imputation in the frequency domain. In particular, we note that the points for both reconstruction ($0\%$ \& $50\%$) and imputation ($50\%$) form a curve (Figure~\ref{fig:freq_corr} Left), suggesting that time series correlation is predictive of frequency correlation.

We additionally visualize the original time series and corresponding spectrogram as well as the CNNAE decoded time series and corresponding spectrogram for both a typical and performant example. A visual inspection shows that our CNNAE produces frequency reconstructions resembling the ground truth spectrograms even for low frequency correlation values. It is possible that alternative metrics for spectrogram evaluation, such as mutual information in frequency space \cite{young2020precise} \cite{malladi2018mutual}, could lead to stronger quantitative correlation metrics compared to the Pearson's Correlation metric that we used for evaluation.

\subsection{Downstream Task: Neural Imputation + Neural Decoding}\label{sec:exp3}
Given neural decoding's significance in the neuroengineering community\cite{carmena2003learning, anumanchipalli2019speech, willett2021high}, we study DNI's ability to directly improve neural decoding performance. We were further motivated by the fact that data used for neural decoding, particularly in humans, is precious and nearly impossible to recreate experimentally if corrupted~\cite{brunton2019data}. We use the movement neural decoder from \cite{peterson2021generalized}, which intakes neural time series data and predicts whether the neural data corresponds to either an arm movement event or rest. After recreating the random forest decoding performance for the original data, we randomly zero-filled either 50\%, 70\%, or 90\% of the data for five random sets of missing electrodes and then computed the resulting performance. We purposefully chose these exaggerated proportions to test DNI's limits on downstream applications with highly corrupted or missing data.

As expected, when large portions of the data are masked via zero-filling, decoding performance decreases significantly. Out of the 36 distinct experiments across 12 participants, our CNNAE-filled data either increased or maintained decoding accuracy in 34/36 experiments (94.44\% of the time). Notably, in the 2 cases where our CNNAE-filled performance dropped below the zero-filled performance (both at 50\%), in the other two experimental categories (70\% and 90\% missing), the CNNAE-filled performance did not drop below the zero-filled performance. These results suggest that there is value in adopting DNI in the neuroscience and neuroengineering communities.  

\begin{figure}
    \centering
        \includegraphics[width=\linewidth]{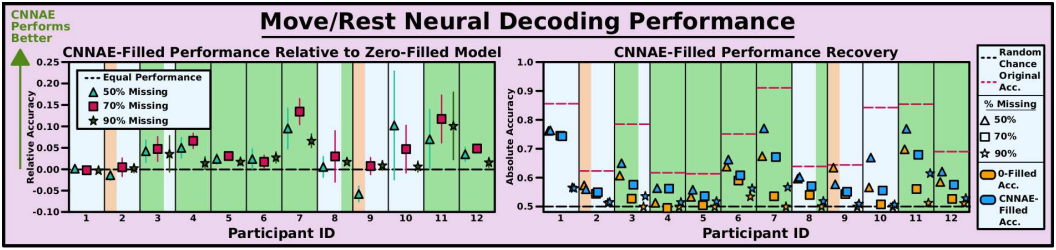}
        \vspace{-0.2in}
        \caption{\textbf{Recovering Neural Decoding Performance When Missing Data is Present}. Background shading indicates (with 1 standard deviation of certainty) when the CNNAE-filled model outperforms (green), performs similarly to (light blue), or underperforms (yellow) compared to the zero-filled model. Each point corresponds to an experimental setting for one participant; results are averaged over 5 runs and the error bars represent 1 standard deviation.}
    \label{fig:diff_recreation}
    \vspace{-0.4cm}
\end{figure}

\section{Related Work}\label{sec:related}
\textbf{Multielectrode Neuroscience Experiments}. Elucidating scientific questions in neuroengineering, neural mechanism discovery, and systems neuroscience has been critically enabled by advances in multielectrode array technology. Different electrode recording modalities (EEG, ECoG, Utah arrays, Neuropixels~\cite{steinmetz2018challenges, paulk2022large}) have specific trade-offs, such as varying signal attenuation or difficulty of implantation, to name a few. However, they all share the commonality that when individual electrodes are corrupted or fail from faulty manufacturing, electrical noise, scar tissue formation, etc., the signal is lost and cannot be experimentally recovered. Previous approaches address neural drift through alignment~\cite{lee2019hierarchical,pandarinath2018latent}, but not neural imputation. In this paper, we recover high-value incomplete brain recordings via our Deep Neural Imputation Framework. 

\textbf{Time Series Imputation}. Our work relates to time series imputation, which aims to recover missing timestamps from time series data. In particular, multivariate time series modeling, where more than one feature is observed at each timestamp, has been studied using statistical methods~\cite{beretta2016nearest,van2000multivariate,acuna2004treatment} as well as deep generative machine learning models~\cite{liu2019naomi,miao2021generative,cao2018brits,che2018recurrent}. Our linear model baseline falls under statistical approaches using nearest-neighbors, while our CNNAE and M-CNNAE approaches are based on generative modeling.  A few common model setups include directly regressing missing values~\cite{cao2018brits,che2018recurrent}, adversarial training (such as GANs)~\cite{miao2021generative,fedus2018maskgan,luo2018multivariate,luo2019e2gan}, as well as autoencoders~\cite{liu2019naomi,luo2019e2gan}. Our CNNAE architecture is based on~\cite{dieleman2021variable}, a model recently developed for time series representation learning, which we adapted for time series imputation via masked electrode modeling. Methods for generic time series imputation typically cannot impute when a feature is missing for all timestamps. These methods are orthogonal to our main contribution, which is a framework for multivariate electrode imputation in both day-generalizable and DOPE-generalizable regimes.

\section{Discussion}\label{sec:discussion}

In multielectrode recordings, signals from corrupted electrodes are commonly discarded and treated as missing. To fill these gaps, we propose Deep Neural Imputation: a framework to recover missing values using neural data across days and individuals. DNI is compatible with linear and nonlinear models, such as deep generative autoencoders, and can be easily be incorporated into existing neuroscience analysis pipelines and downstream tasks.

\textbf{Key Observations.} We find that our nearest neighbor linear baseline reconstructs and imputes missing electrode data well when there is a small percentage of data missing (e.g. $10\%$). Further, both the CNNAE and M-CNNAE models recover neural data in day-generalizable and DOPE-generalizable manners respectively when there are large percentages of data missing (e.g. $50\%$). These results may inspire neuroscientists to adjust neural electrode placement or enable novel experimental design. Moreover, there is a positive relationship between frequency correlation and time-series correlation, indicating that without training explicitly for frequency content, we can reconstruct and impute it. Finally, DNI is practical in downstream neuroscience tasks, which we validate based on significant improvement on a neural decoding task in the presence of missing data.

\textbf{Limitations and Future Directions.} There are many directions for future work stemming from the neural imputation problem. For joint-participant experiments, our method currently requires a participant-specific layer to be trained for each spatial configuration. Further explorations into transfer learning and meta-learning could eliminate this requirement. The joint-participant model offers an intriguing possibility of studying the joint embedding of neural representations to, for instance, understand individual variability in neural correlates of similar behaviors~\cite{peterson2021behavioral}. Additionally, our models cannot currently impute data gathered from unobserved spatial locations. Graph neural networks are one possible architecture that could develop this capability. Furthermore, these models may be improved by exploring multimodal fusion of neural, kinematic, and other measurement modalities~\cite{barnum2020benefits,peterson2021learning}. Since no prior work has conducted a rigorous study of how to recover missing electrode recordings across days and participants (see an early attempt~\cite{wang2018brains}), establishing multi-participant, multi-day benchmark of DNI methods would be valuable.

\textbf{Significance and Broader impacts}.
Participants' neural activity is deeply personal; therefore, we need to be thoughtful in our use of technology to analyze this information. The ethical considerations include issues of data management to guard the subjects' privacy and security.  At the same time, we need to consider this technology's assistive potential in improving participants' lives, because recovering missing data from incomplete brain recordings has the potential to transform the design and analysis of a wide range of neuroscience and neuroengineering studies. Further discussion and information on the data collection process can be found in the original dataset release paper~\cite{peterson2022ajile12}. 

\section{Acknowledgements}

We thank Albert Hao Li for thoughtful discussions and feedback throughout the project, Steve Peterson \& Zoe Steine-Hanson for sharing their AJILE12 dataset knowledge, and Ann Kennedy for helpful conversations. This work was supported by an NSF Graduate Fellowship (to ST), NSERC Award \#PGSD3-532647-2019 (to JJS), and the Moore Distinguished Scholar Program at Caltech (to BWB).

\newpage

\setcitestyle{numbers}
\bibliographystyle{unsrt}
\bibliography{refs}

\newpage

\appendix

\newpage
\appendix

\section*{Appendix}

\section{Additional Experimental Results}~\label{sec:appendix_res}
We present additional experimental results for experiments in Sections 4.3, 4.4, and 4.5 in Appendix Sections~\ref{sec:a1}, ~\ref{sec:a2}, and ~\ref{sec:a3} respectively.

\subsection{Benchmark Results with Standard Deviation}~\label{sec:a1}
The mean and standard deviation from the correlation comparison in Figure 4 of the main paper is shown in Tables~\ref{table:results}, ~\ref{table:results2}. At $0\%$ missing electrodes, both the CNNAE and M-CNNAE generally outperform the baseline. In addition, the M-CNNAE performs better than the CNNAE for participant 1, and similarly to the CNNAE for the other participants. The gap between the machine learning models and the baseline increases as the amount of missing electrode increases - in particular, there is a significant improvement using the CNNAE and M-CNNAE across all participants at $50\%$ of missing electrodes. Similar to Figure 4, the CNNAE and M-CNNAE generally has higher correlation compared to the baseline at imputation as well as reconstruction.

\subsection{Frequency Analysis for All Participants}~\label{sec:a2}
\begin{figure}
    \centering
        \includegraphics[width=\linewidth]{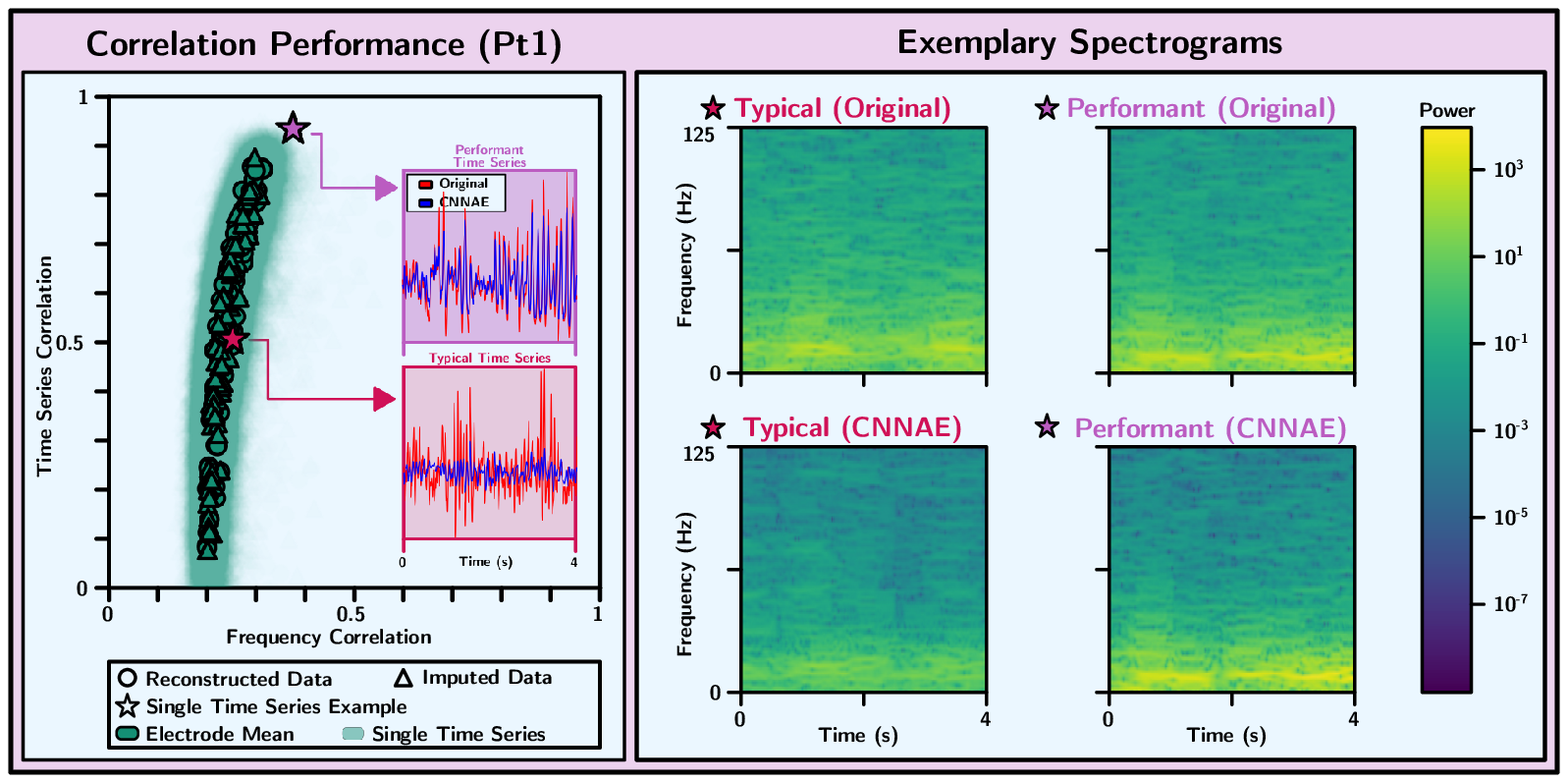}
        \vspace{-0.2in}
        \caption{\textbf{Relationship between Frequency and Time Series Correlation}. \textit{Left:} We show the frequency and time-series correlation from the CNNAE for participant 1, where each outlined point corresponds to one electrode. Two samples of reconstructed time series relative to the original are depicted. \textit{Right:} Spectrogram examples corresponding to the time series examples on the left for a typical example and a performant example. }
    \label{fig:freq_corr_pt1}
    \vspace{-0.25cm}
\end{figure}

\begin{figure}
    \centering
        \includegraphics[width=\linewidth]{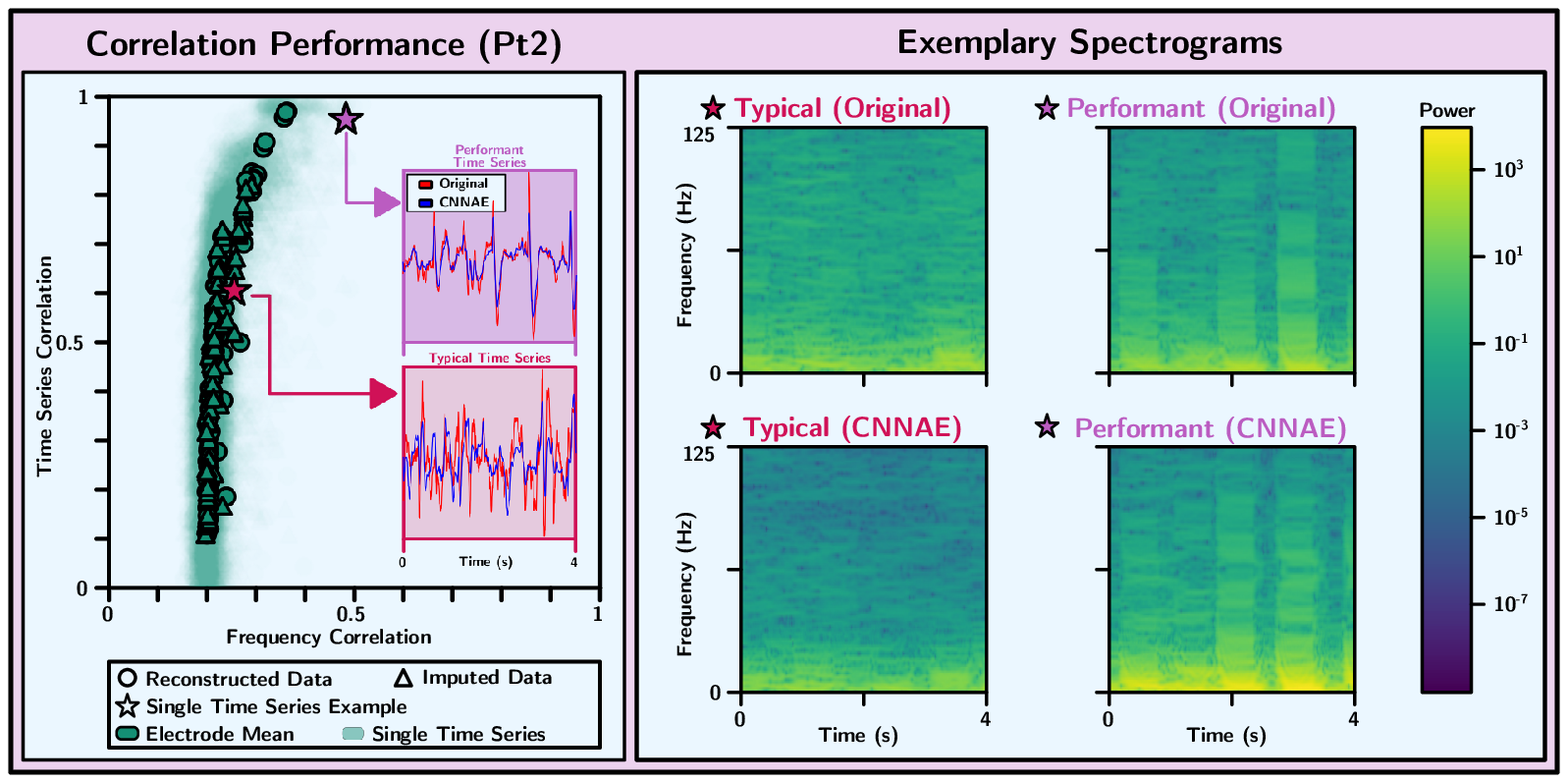}
        \vspace{-0.2in}
        \caption{\textbf{Relationship between Frequency and Time Series Correlation}. \textit{Left:} We show the frequency and time-series correlation from the CNNAE for participant 2, where each outlined point corresponds to one electrode. Two samples of reconstructed time series relative to the original are depicted. \textit{Right:} Spectrogram examples corresponding to the time series examples on the left for a typical example and a performant example. }
    \label{fig:freq_corr_pt2}
    \vspace{-0.25cm}
\end{figure}

\begin{figure}
    \centering
        \includegraphics[width=\linewidth]{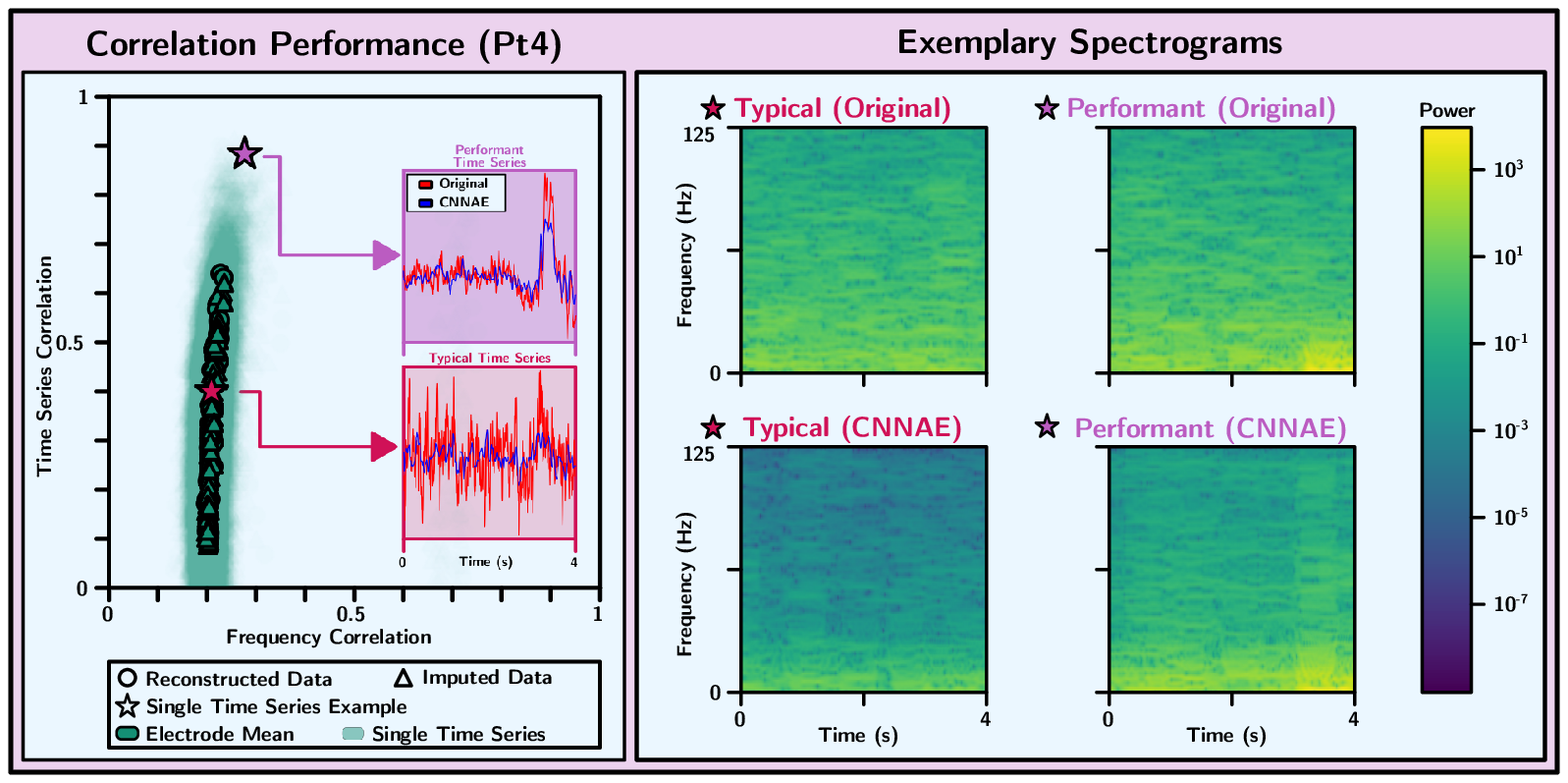}
        \vspace{-0.2in}
        \caption{\textbf{Relationship between Frequency and Time Series Correlation}. \textit{Left:} We show the frequency and time-series correlation from the CNNAE for participant 4, where each outlined point corresponds to one electrode. Two samples of reconstructed time series relative to the original are depicted. \textit{Right:} Spectrogram examples corresponding to the time series examples on the left for a typical example and a performant example. }
    \label{fig:freq_corr_pt4}
    \vspace{-0.25cm}
\end{figure}

\begin{figure}
    \centering
        \includegraphics[width=\linewidth]{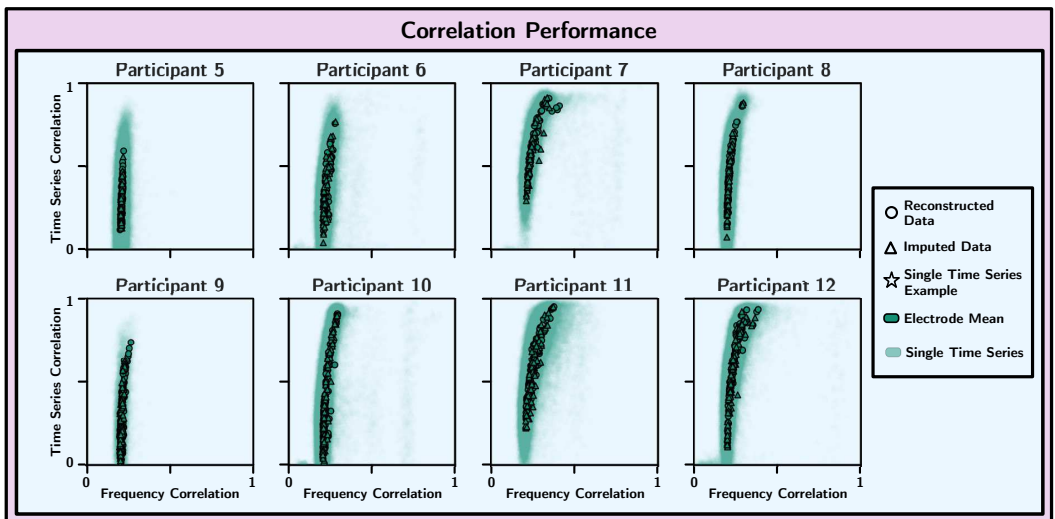}
        \vspace{-0.2in}
        \caption{\textbf{Relationship between Frequency and Time Series Correlation}. \textit{Left:} We show the frequency and time-series correlation from the CNNAE for participant 5, 6, 7, 8, 9, 10, 11, 12, where each outlined point corresponds to one electrode.}
    \label{fig:freq_corr_rest_patients}
    \vspace{-0.25cm}
\end{figure}

Here we expand on the participant 3 results presented in the paper, Figure \ref{fig:freq_corr}, by showing the results for all participants. For participants 1, 2, 4 we show not only the time series and frequency correlation plots but also some example time series and spectrograms (Figures \ref{fig:freq_corr_pt1}, \ref{fig:freq_corr_pt2}, \ref{fig:freq_corr_pt4} respectively). In addition, for the other participants 5, 6, 7, 8, 9, 10, 11, 12 we show the time series and frequency correlation plots, Figure \ref{fig:freq_corr_rest_patients}. Results for all patients are shown across two proportions of missing data, 10\% and 50\%.

We observed that there was a positive relationship between time series correlation and frequency correlation, despite not having trained our model on frequency reconstruction. This trend between time series correlation and frequency correlation can been seen strongest in participants 1, 2, 3, 6, 7, 8, 10, 11, and 12.

\subsection{Neural Decoding Results}~\label{sec:a3}
Table \ref{table:results_neural_decoding} expands on the results presented in Figure \ref{fig:diff_recreation}. In Table \ref{table:results_neural_decoding}, we include the performance of the random forest neural decoder across 50\%, 70\%, and 90\% of electrodes missing. We report the mean performance across 5 random seeds for the zero-filled data, and CNNAE-filled data. In addition, we include the mean and standard deviation for the pairwise relative accuracy between the CNNAE-filled data and zero-filled data. Positive mean relative accuracy values indicate that the CNNAE-filled data outperforms the zero-filled data on the move/rest neural decoding task.

\begin{table*}
\footnotesize
\centering
\begin{tabular}{|c|c||*{4}{c|}}
    \hline
    \multicolumn{2}{|c||}{}\bigstrut & \multicolumn{4}{|c|}{\% Electrodes Missing} \\
    \cline{3-6}
    \multicolumn{2}{|c||}{}\bigstrut[t] & 0\% & 10\% & 20\% & 50\% \\
    \hline
    
    \hline
    \multirow{6}{*}{Pt 1}\bigstrut[t] & \multirow{2}{*}{Baseline} & $.536$ & $.503 \pm .007$ & $.486 \pm .012$ & $.375 \pm .027$ \\
     & & - & $.585 \pm .010$ & $.480 \pm .023$ & $.356 \pm .020$ \\
     \cline{2-6}
    & \multirow{2}{*}{\textbf{CNNAE}} & $.521 \pm .005$ & $.516 \pm .005$ & $.519 \pm .011$ & $.542 \pm .014$ \\
    & & - & $.497 \pm .033$ & $.461 \pm .024$ & $.411 \pm .016$ \\
    \cline{2-6}
    & \multirow{2}{*}{\textbf{M-CNNAE}} & $.569 \pm .008$ & $.565 \pm .009$ & $.566 \pm .015$ & $.579 \pm .019$ \\
    & & - & $.534 \pm .034$ & $.492 \pm .021$ & $.434 \pm .023$\\
    
    \hline
    \multirow{6}{*}{Pt 2}\bigstrut[t] & \multirow{2}{*}{Baseline} & $.662$ & $.641 \pm .006$ & $.624 \pm .014$ & $.484 \pm .031$ \\
     & & - & $.661 \pm .067$ & $.604 \pm .045$ & $.488 \pm .031$ \\
     \cline{2-6}
    & \multirow{2}{*}{\textbf{CNNAE}} & $.700 \pm .002$ & $.691 \pm .005$ & $.690 \pm .005$ & $.653 \pm .013$ \\
    & & - & $.731 \pm .048$ & $.688 \pm .0029$ & $.674 \pm .014$ \\
    \cline{2-6}
    & \multirow{2}{*}{\textbf{M-CNNAE}} & $.710 \pm .007$ & $.700 \pm .008$ & $.697 \pm .008$ & $.645 \pm .014$ \\
    & & - & $.741 \pm .053$ & $.696 \pm .033$ & $.667 \pm .012$\\
    
    \hline
    \multirow{6}{*}{Pt 3}\bigstrut[t] & \multirow{2}{*}{Baseline} & $.622$ & $.583 \pm .013$ & $.545 \pm .006$ & $.448 \pm .031$ \\
     & & - & $.677 \pm .009$ & $.619 \pm .038$ & $.425 \pm .033$ \\
    \cline{2-6}
    & \multirow{2}{*}{\textbf{CNNAE}} & $.621 \pm .001$ & $.609 \pm .003$ & $.604 \pm .003$ & $.617 \pm .018$ \\
    & & - & $.682 \pm .010$ & $.634 \pm .019$ & $.510 \pm .022$ \\
    \cline{2-6}
    & \multirow{2}{*}{\textbf{M-CNNAE}} & $.632 \pm .008$ & $.620 \pm .009$ & $.615 \pm .010$ & $.624 \pm .021$ \\
    & & - & $.679 \pm .010$ & $.634 \pm .019$ & $.503 \pm .021$\\
    
    \hline
    \multirow{6}{*}{Pt 4}\bigstrut[t] & \multirow{2}{*}{Baseline} & $.451$ & $.419 \pm .007$ & $.396 \pm .016$ & $.294 \pm .011$ \\
     & & - & $.505 \pm .044$ & $.464 \pm .05$ & $.327 \pm .011$ \\
     \cline{2-6}
    & \multirow{2}{*}{\textbf{CNNAE}} & $.559 \pm .003$ & $.551 \pm .007$ & $.551 \pm .013$ & $.534 \pm .014$ \\
    & & - & $.573 \pm .035$ & $.524 \pm .046$ & $.466 \pm .009$ \\
    \cline{2-6}
    & \multirow{2}{*}{\textbf{M-CNNAE}} & $.550 \pm .008$ & $.543 \pm .011$ & $.541 \pm .015$ & $.516 \pm .010$ \\
    & & - & $.565 \pm .039$ & $.509 \pm .041$ & $.458 \pm .011$\\
    
    \hline
    \multirow{6}{*}{Pt 5}\bigstrut[t] & \multirow{2}{*}{Baseline} & $.549$ & $.532 \pm .004$ & $.500 \pm .015$ & $.402 \pm .015$ \\
     & & - & $.494 \pm .042$ & $.475 \pm .035$ & $.376 \pm .025$ \\
     \cline{2-6}
    & \multirow{2}{*}{\textbf{CNNAE}} & $.581 \pm .012$ & $.577 \pm .014$ & $.574 \pm .016$ & $.556 \pm .021$ \\
    & & - & $.543 \pm .035$ & $.540 \pm .035$ & $.518 \pm .017$ \\
    \cline{2-6}
    & \multirow{2}{*}{\textbf{M-CNNAE}} & $.573 \pm .006$ & $.573 \pm .007$ & $.570 \pm .011$ & $.550 \pm .015$ \\
    & & - & $.538 \pm .039$ & $.539 \pm .034$ & $.514 \pm .014$\\
    
    \hline
    \multirow{6}{*}{Pt 6}\bigstrut[t] & \multirow{2}{*}{Baseline} & $.453$ & $.436 \pm .002$ & $.408 \pm .009$ & $.274 \pm .024$ \\
     & & - & $.425 \pm .025$ & $.411 \pm .020$ & $.311 \pm .031$ \\
     \cline{2-6}
    & \multirow{2}{*}{\textbf{CNNAE}} & $.483 \pm .013$ & $.483 \pm .017$ & $.487 \pm .016$ & $.492 \pm .019$ \\
    & & - & $.420 \pm .019$ & $.406 \pm .029$ & $.366 \pm .016$ \\
    \cline{2-6}
    & \multirow{2}{*}{\textbf{M-CNNAE}} & $.499 \pm .007$ & $.500 \pm .008$ & $.502 \pm .007$ & $.495 \pm .015$ \\
    & & - & $.421 \pm .009$ & $.408 \pm .024$ & $.364 \pm .007$\\
    
    \hline
\end{tabular}
\caption{\textbf{Correlation of Deep Neural Imputation Methods.} We show the time series correlation for the linear model baseline, CNNAE, and M-CNNAE. The top row for each method represents reconstruction, and the bottom row represents imputation. The mean and standard deviation is from 9 runs (3 runs of each model with 3 sets of missing data). Note that virtually all prior methods ignore missing data, therefore the conventional baseline approach would get 0 correlation. Participants 1 to 6 is shown, and due to space, participants 7 to 12 is in Table~\ref{table:results2}.}
\label{table:results}
\end{table*}

\begin{table*}
\footnotesize
\centering
\begin{tabular}{|c|c||*{4}{c|}}
    \hline
    \multicolumn{2}{|c||}{}\bigstrut & \multicolumn{4}{|c|}{\% Electrodes Missing} \\
    \cline{3-6}
    \multicolumn{2}{|c||}{}\bigstrut[t] & 0\% & 10\% & 20\% & 50\% \\
    \hline
    
    \hline

    \multirow{6}{*}{Pt 7}\bigstrut[t] & \multirow{2}{*}{Baseline} & $.577$ & $.561 \pm .007$ & $.540 \pm .015$ & $.373 \pm .032$ \\
     & & - & $.487 \pm .052$ & $.465 \pm .023$ & $.395 \pm .016$ \\
     \cline{2-6}
    & \multirow{2}{*}{\textbf{CNNAE}} & $.639 \pm .016$ & $.659 \pm .007$ & $.658 \pm .006$ & $.611 \pm .016$ \\
    & & - & $.531 \pm .063$ & $.564 \pm .011$ & $.570 \pm .027$ \\
    \cline{2-6}
    & \multirow{2}{*}{\textbf{M-CNNAE}} & $.664 \pm .010$ & $.668 \pm .012$ & $.662 \pm .011$ & $.616 \pm .015$ \\
    & & - & $.553 \pm .068$ & $.584 \pm .008$ & $.569 \pm .025$\\    
    \hline
    
    \multirow{6}{*}{Pt 8}\bigstrut[t] & \multirow{2}{*}{Baseline} & $.483$ & $.461 \pm .006$ & $.444 \pm .019$ & $.332 \pm .026$ \\
     & & - & $.478 \pm .024$ & $.418 \pm .047$ & $.313 \pm .045$ \\
     \cline{2-6}
    & \multirow{2}{*}{\textbf{CNNAE}} & $.588 \pm .002$ & $.590 \pm .002$ & $.592 \pm .005$ & $.575 \pm .028$ \\
    & & - & $.511 \pm .016$ & $.509 \pm .016$ & $.476 \pm .005$ \\
    \cline{2-6}
    & \multirow{2}{*}{\textbf{M-CNNAE}} & $.570 \pm .007$ & $.569 \pm .006$ & $.569 \pm .008$ & $.535 \pm .028$ \\
    & & - & $.487 \pm .025$ & $.480 \pm .016$ & $.454 \pm .005$\\
    
    \hline
    \multirow{6}{*}{Pt 9}\bigstrut[t] & \multirow{2}{*}{Baseline} & $.637$ & $.616 \pm .006$ & $.566 \pm .011$ & $.436 \pm .038$ \\
     & & - & $.608 \pm .080$ & $.646 \pm .054$ & $.457 \pm .013$ \\
     \cline{2-6}
    & \multirow{2}{*}{\textbf{CNNAE}} & $.582 \pm .001$ & $.585 \pm .006$ & $.573 \pm .009$ & $.532 \pm .016$ \\
    & & - & $.533 \pm .054$ & $.587 \pm .044$ & $.568 \pm .029$  \\
    \cline{2-6}
    & \multirow{2}{*}{\textbf{M-CNNAE}} & $.583 \pm .006$ & $.584 \pm .008$ & $.570 \pm .012$ & $.521 \pm .011$ \\
    & & - & $.537 \pm .056$ & $.584 \pm .045$ & $.557 \pm .032$\\
    
    \hline
    \multirow{6}{*}{Pt 10}\bigstrut[t] & \multirow{2}{*}{Baseline} & $.552$ & $.529 \pm .015$ & $.508 \pm .020$ & $.406 \pm .015$ \\
     & & - & $.480 \pm .110$ & $.490 \pm .076$ & $.378 \pm .014$ \\
     \cline{2-6}
    & \multirow{2}{*}{\textbf{CNNAE}} & $.526 \pm .002$ & $.529 \pm .004$ & $.529 \pm .008$ & $.501 \pm .014$ \\
    & & - & $.480 \pm .020$ & $.490 \pm .018$ & $.503 \pm .019$ \\
    \cline{2-6}
    & \multirow{2}{*}{\textbf{M-CNNAE}} & $.532 \pm .006$ & $.533 \pm .006$ & $.532 \pm .010$ & $.502 \pm .018$ \\
    & & - & $.483 \pm .024$ & $.494 \pm .019$ & $.506 \pm .020$\\
    
    \hline
    \multirow{6}{*}{Pt 11}\bigstrut[t] & \multirow{2}{*}{Baseline} & $.617$ & $.590 \pm .004$ & $.553 \pm .007$ & $.444 \pm .074$ \\
     & & - & $.594 \pm .028$ & $.575 \pm .038$ & $.401 \pm .054$ \\
     \cline{2-6}
    & \multirow{2}{*}{\textbf{CNNAE}} & $.701 \pm .001$ & $.699 \pm .005$ & $.700 \pm .005$ & $.700 \pm .024$ \\
    & & - & $.671 \pm .044$ & $.661 \pm .018$ & $.613 \pm .024$ \\
    \cline{2-6}
    & \multirow{2}{*}{\textbf{M-CNNAE}} & $.678 \pm .009$ & $.676 \pm .011$ & $.675 \pm .011$ & $.669 \pm .024$ \\
    & & - & $.648 \pm .047$ & $.636 \pm .020$ & $.587 \pm .025$\\
    
    \hline
    \multirow{6}{*}{Pt 12}\bigstrut[t] & \multirow{2}{*}{Baseline} & $.485$ & $.461 \pm .005$ & $.442 \pm .008$ & $.352 \pm .031$ \\
     & & - & $.496 \pm .031$ & $.427 \pm .015$ & $.33 \pm .023$ \\
     \cline{2-6}
    & \multirow{2}{*}{\textbf{CNNAE}} & $.549 \pm .002$ & $.536 \pm .002$ & $.535 \pm .006$ & $.515 \pm .008$ \\
    & & - & $.600 \pm .027$ & $.547 \pm .018$ & $.507 \pm .009$ \\
    \cline{2-6}
    & \multirow{2}{*}{\textbf{M-CNNAE}} & $.535 \pm .007$ & $.522 \pm .007$ & $.520 \pm .009$ & $.494 \pm .013$ \\
    & & - & $.585 \pm .027$ & $.532 \pm .019$ & $.479 \pm .011$\\
    \hline
\end{tabular}
\caption{\textbf{Correlation of Deep Neural Imputation Methods.} We show the time series correlation for the linear model baseline, CNNAE, and M-CNNAE. The top row for each method represents reconstruction, and the bottom row represents imputation. The mean and standard deviation is from 9 runs (3 runs of each model with 3 sets of missing data). Note that virtually all prior methods ignore missing data, therefore the conventional baseline approach would get 0 correlation. Participants 7 to 12 is shown, and due to space, participants 1 to 6 is in Table~\ref{table:results}.}
\label{table:results2}
\end{table*}

\begin{table*}
\footnotesize
\centering
\begin{tabular}{|c|c||*{3}{c|}}
    \hline
    \multicolumn{2}{|c||}{}\bigstrut & \multicolumn{3}{|c|}{\% Electrodes Missing} \\
    \cline{3-5}
    \multicolumn{2}{|c||}{}\bigstrut[t] & 50\% & 70\% & 90\% \\
    \hline
    
    \hline
    \multirow{3}{*}{Pt 1}\bigstrut[t] & Zero-Filled & $0.7597$ & $0.7430$ & $0.5652$ \\
    & CNNAE-Filled & $0.7615$ & $0.7412$ & $0.5633$ \\
    & Pairwise Relative Accuracy & $0.0018\pm 0.0034$ & $-0.0018\pm 0.0044$ & $-0.0020\pm0.0039$ \\
    \hline
    \multirow{3}{*}{Pt 2}\bigstrut[t] & Zero-Filled & $0.5726$ & $0.5437$ & $0.5126$ \\
    & CNNAE-Filled & $0.5593$ & $0.5489$ & $0.5156$ \\
    & Pairwise Relative Accuracy & $-0.0133\pm 0.0127$ & $0.0052 \pm 0.0217$ & $0.0030\pm 0.0108$ \\
    \hline
    \multirow{3}{*}{Pt 3}\bigstrut[t] & Zero-Filled & $0.6062$ & $0.5271$ & $0.4997$ \\
    & CNNAE-Filled & $0.6487$ & $0.5748$ & $0.5359$ \\
    & Pairwise Relative Accuracy & $0.0425\pm 0.0266$ & $0.0477\pm 0.0291$ & $0.0363 \pm 0.0426$ \\
    \hline
    \multirow{3}{*}{Pt 4}\bigstrut[t] & Zero-Filled & $0.5128$ & $0.4948$ & $0.5000$ \\
    & CNNAE-Filled & $0.5629$ & $0.5614$ & $0.5151$ \\
    & Pairwise Relative Accuracy & $0.0501\pm 0.0240$ & $0.0667 \pm 0.0181$ & $0.0151 \pm 0.0149$ \\
    \hline    
    \multirow{3}{*}{Pt 5}\bigstrut[t] & Zero-Filled & $0.5331$ & $0.5044$ & $0.5000$ \\
    & CNNAE-Filled & $0.5576$ & $0.5357$ & $0.5181$ \\
    & Pairwise Relative Accuracy & $0.0245\pm 0.0130$ & $0.0313 \pm 0.0028$ & $0.0181 \pm 0.0088$ \\
    \hline
    \multirow{3}{*}{Pt 6}\bigstrut[t] & Zero-Filled & $0.6370$ & $0.5893$ & $0.5343$ \\
    & CNNAE-Filled & $0.6613$ & $0.6070$ & $0.5624$ \\
    & Pairwise Relative Accuracy & $0.0243\pm 0.0237$ & $0.0178 \pm 0.0156$ & $0.0281 \pm 0.0133$ \\
    \hline
    \multirow{3}{*}{Pt 7}\bigstrut[t] & Zero-Filled & $0.6728$ & $0.5353$ & $0.5000$ \\
    & CNNAE-Filled & $0.7682$ & $0.6699$ & $0.5669$ \\
    & Pairwise Relative Accuracy & $0.0954\pm 0.0477$ & $0.1347 \pm 0.0305$ & $0.0669 \pm 0.0163$ \\
    \hline
    \multirow{3}{*}{Pt 8}\bigstrut[t] & Zero-Filled & $0.5957$ & $0.5395$ & $0.5009$ \\
    & CNNAE-Filled & $0.6015$ & $0.5699$ & $0.5187$ \\
    & Pairwise Relative Accuracy & $0.0058\pm 0.0242$ & $0.0303\pm 0.0600$ & $0.0178 \pm 0.0047$ \\
    \hline
    \multirow{3}{*}{Pt 9}\bigstrut[t] & Zero-Filled & $0.6333$ & $0.5429$ & $0.5000$ \\
    & CNNAE-Filled & $0.5762$ & $0.5508$ & $0.5095$ \\
    & Pairwise Relative Accuracy & $-0.0571\pm 0.0184$ & $0.0079 \pm 0.0201$ & $0.0095 \pm 0.0117$ \\
    \hline
    \multirow{3}{*}{Pt 10}\bigstrut[t] & Zero-Filled & $0.5656$ & $0.5074$ & $0.5000$ \\
    & CNNAE-Filled & $0.6680$ & $0.5548$ & $0.5069$ \\
    & Pairwise Relative Accuracy & $0.1024\pm 0.1269$ & $0.0474 \pm 0.0557$ & $0.0069 \pm 0.0131$ \\
    \hline
    \multirow{3}{*}{Pt 11}\bigstrut[t] & Zero-Filled & $0.6966$ & $0.5601$ & $0.5127$ \\
    & CNNAE-Filled & $0.7668$ & $0.6777$ & $0.6141$ \\
    & Pairwise Relative Accuracy & $0.0703\pm 0.0697$ & $0.1176 \pm 0.0559$ & $0.1014 \pm 0.0789$ \\
    \hline
    \multirow{3}{*}{Pt 12}\bigstrut[t] & Zero-Filled & $0.5845$ & $0.5262$ & $0.5120$ \\
    & CNNAE-Filled & $0.6199$ & $0.5749$ & $0.5286$ \\
    & Pairwise Relative Accuracy & $0.0354\pm 0.0165$ & $0.0487\pm 0.0084$ & $0.0166 \pm 0.0120$ \\
    \hline
\end{tabular}
\caption{\textbf{CNNAE Recovers Neural Decoding Performance} (corresponding to Figure 6). Move/rest neural decoding performance absolute means, and pairwise relative accuracy means \& standard deviations. The results for each percent missing (50\%, 70\%, and 90\%) were calculated across 5 seeds. There are only two instances where the zero-filled accuracy outperformed the CNNAE-filled accuracy, corresponding to the yellow shading in Figure 6. All other cases correspond to where the CNNAE-filled accuracy was equal to or outperformed the zero-filled accuracy.}
\label{table:results_neural_decoding}
\end{table*}

\section{Implementation Details}~\label{sec:appendix_implementation}
We provide additional details on the data processing and the model implementation. Model hyperparameters are in Table~\ref{tab:hyperparameter_1}.


\subsection{Data Processing}
We will start by providing more intuition behind the type of neural time series that is collected from the participants. Each electrode in the ECoG array that participants are implanted with generates a local field potential (LFP) voltage trace. Each LFP trace comes from the bulk voltage activity of a few tens-of-thousands to hundreds-of-thousands of neurons in the brain. These LFP voltage traces the comprise the AJILE12 dataset are stored in the common neuroscience data file format, NWB, which is specifically designed to handle ragged data, missing values, and non-task-dependent (i.e. naturalistic) neural data \cite{ruebel2019nwb}.

In the AJILE12 dataset there are 12 human participants; each participant has roughly 100 ECoG electrodes implanted in their brain based on clinically determined locations. Before pre-processing, the data is collected at 500Hz and in most cases continuously recorded for several days. Given the multitude of goals in our paper we created two data processing pipelines to explore our models: Procedure A, and Procedure B. In Procedure A, we standardize 50,000 sample segments of the AJILE12 dataset and downsample the data to 5Hz to study our ability to perform time series reconstruction. In Procedure B, we following the data processing pipeline in \cite{peterson2021generalized}, producing 250Hz data that allows us to study our frequency reconstruction ability, and our model's capability at restoring move/rest neural decoding performance.

\subsection{Model Details}

\textbf{Linear Baseline Model}. In our linear baseline, we first break the patients' recording days into "train" days and a "test" day. This data split follows the dataset split in \cite{peterson2021generalized}, where the last day is held out as the test day. We then use the full set of electrodes for each patient to create a distance matrix which allows us to calculate each electrode's nearest neighbors by euclidean distance. The main idea is to linearly combine the time series correlation of the nearest neighbors to impute missing electrodes.

For every electrode on the test day, we compute the time series correlations between the target electrode and its three nearest neighbors averaged across observed training days. Then on the test day, the nearest neighbors are combined linearly using the train day correlations. Step-by-step, this corresponds to: (1) select a test day target electrode, (2) find one of its three nearest neighbors, (3) get the previously calculated time series correlation value that was averaged across all training days and all time series for that target electrode and the specific nearest neighbor, (4) take the time series from this nearest neighbor on the test day and weight the time series (multiply it) by the training time series correlation, (5) repeat 2-4 for all 3 nearest neighbors [if a neighbor is not observed on the test day, we use only the correlation from observed neighbors], (6) take the sum of the three weighted time series as the baseline reconstruction. 

For evaluation, we (1) calculate the correlation between this baseline reconstruction and the time series of the actual target electrode recordings, (2) after calculating the correlations for all the time series off of an electrode we can average across all our time series to get the per electrode correlation.

\textbf{CNNAE}. The CNNAE model is trained for each participant. The model architecture is based on the encoder-decoder setup in~\cite{dieleman2021variable}, with strided temporal convolutions for encoding layers, an upsampling network, and a WaveNet decoder~\cite{oord2016wavenet}. Our model hyperparameters are not tuned to AJILE12 and doing a hyperparameter search on the validation set can likely further improve performance. The model is trained to convergence of the train loss at epoch 100. 

We input the time series data for the full set of electrodes for each participant (with missing electrodes zero-filled) to the model. At the input layer, we input both the electrode time series data as well as the time derivative of the data (change in time series value across 1 timestamp). The encoder downsamples the time series by a factor of 8 with the current kernel and stride configurations~\ref{tab:hyperparameter_1}. We then upsample the encoding by a factor of 8 using the upsampling network before input to the decoder. The WaveNet decoder preserves the sequence length, and the output of WaveNet is fed into the final shallow temporal convolution layers to output the electrode reconstruction and imputation. There are four sets of shallow networks, outputting the mean and variance of the time series data and the time derivative. During evaluation, we use the output mean of the time series data.

Our CNNAE models are trained on either Amazon EC2 using p2 instances with a single Tesla K80 GPU, or a single NVIDIA GeForce RTX 2080Ti GPU. In addition to maximizing the log-likelihood of reconstruction and imputation during training, we also use the margin loss and slowness penalty proposed in~\cite{dieleman2021variable} to regularize the embeddings.

\textbf{M-CNNAE}. The M-CNNAE model is trained jointly for all participants, with a similar architecture to the CNNAE. Our M-CNNAE model hyperparameters are not tuned to AJILE12 and doing a hyperparameter search on the validation set can likely further improve performance. The model is trained to epoch 45. The main difference for the M-CNNAE compared to the CNNAE is participant-specific encoding and decoding layers, with shared layers across participants similar to the CNNAE in between. 

Given input time series data, the data is mapped by a participant-specific encoding layer before feeding into the CNNAE encoder, which is a set of strided convolutions. Then similar to the CNNAE, the encodings are upsampled, then fed into the WaveNet decoder. The decoder outputs are used as inputs to the final participant-specific decoding layers, which are shallow networks that map decoder outputs to the mean and variance of the time series data and the time derivative. During evaluation, we use the output mean of the time series data.

Our M-CNNAE models are trained on Amazon EC2 using p2 instances with a single Tesla K80 GPU. In addition to maximizing the log-likelihood of reconstruction and imputation during training, we also use the margin loss and slowness penalty proposed in~\cite{dieleman2021variable} to regularize the embeddings.

\begin{table}[!t]
  \centering
  \small
  \scalebox{1.0}{
   \begin{tabular}{| c | c | c| c | c| c | c |c |} 
   \hline
   Model & Batch size & Learning Rate & z-dim & Upsampling & Num units & Encoder & Decoder \\
   \hline
    CNNAE & 16 & 0.0001 & 64 & 8x & 256 & Kernel: [4,4,4] & Layers: 2 \\
     &  &  &  &  &  & Stride: [2,2,2] & Blocks: 2 \\    
    \hline
    M-CNNAE & 2 & 0.0001 & 64 & 8x & 256 & Kernel: [4,4,4] & Layers: 2 \\
     & (per pt) &  &  &  &  & Stride: [2,2,2] & Blocks: 2 \\
   \hline
\end{tabular}
}
\vspace{0.1in}
\caption{{\bf Hyperparameters for CNNAE and M-CNNAE.} The encoder is a set of strided convolutions with the specific kernel and stride parameters at each layer. The decoder is based on WaveNet~\cite{oord2016wavenet}. The number of units is for both the encoder and decoder. The M-CNNAE has the same architecture as the CNNAE, except for shallow participant-specific encoding and decoding layers at the input and output respectively. } \label{tab:hyperparameter_1}
\end{table}


\end{document}